\documentclass{esannV2}

\usepackage[latin1]{inputenc}
\usepackage{amssymb,amsmath,array}

\usepackage{hyperref}
\usepackage[pdftex]{graphicx}

\usepackage{multirow}
\usepackage{algorithm,algpseudocode,float}

\usepackage{booktabs}

\usepackage{amsmath,amssymb,amsfonts}
\usepackage{graphicx}
\usepackage{subfig} 

\begin{document}

\title{Continual Contrastive  Learning on Tabular Data with Out of Distribution}

\author{Achmad Ginanjar$^{a}$
\thanks{Indonesia Endowment Fund for Education Agency (LPDP)}
, Xue Li$^a$, Priyanka Singh$^a$, and Wen Hua$^b$  
\vspace{.3cm}\\
$^a$ School of Electrical Engineering and Computer Science\\
The University of Queensland,
Queensland, Australia\\
$^b$ Department of Computing, The Hong Kong Polytechnic University 
Hong Kong\\
}

\maketitle

\begin{abstract}
Out-of-distribution (OOD) prediction remains a significant challenge in machine learning, particularly for tabular data where traditional methods often fail to generalize beyond their training distribution. This paper introduces Tabular Continual Contrastive Learning (TCCL), a novel framework designed to address OOD challenges in tabular data processing. TCCL integrates contrastive learning principles with continual learning mechanisms, featuring a three-component architecture: an Encoder for data transformation, a Decoder for representation learning, and a Learner Head. We evaluate TCCL against 14 baseline models, including state-of-the-art deep learning approaches and gradient-boosted decision trees (GBDT), across eight diverse tabular datasets. Our experimental results demonstrate that TCCL consistently outperforms existing methods in both classification and regression tasks on OOD data, with particular strength in handling distribution shifts. These findings suggest that TCCL represents a significant advancement in handling OOD scenarios for tabular data.
\end{abstract}

\section{Introduction}
When a machine learning model encounters new data that is out of distribution (OOD), it may experience a significant decline in performance \cite{Hsu2020}. OOD refers to data samples that differ from the distribution of the training data, which can often lead to unreliable predictions \cite{OODBaseline}. Addressing the challenges associated with OOD is crucial for maintaining a model's performance. This highlights the significance of our research in identifying effective strategies for managing OOD scenarios.

Significant progress has been made in OOD detection using algorithms such as MCDD \cite{deepMCCD}, OpenMax \cite{Bendale2016}, and TemperatureScaling \cite{platt1999probabilistic}. However, challenges remain for prediction tasks involving tabular data. Recent advancements in deep learning for tabular data show promise \cite{subtab,tabr,scarf}; however, Gradient Boosted Decision Trees (GBDT) still tend to be the best option for tabular datasets \cite{Grinsztajn2022}. On the other hand, the  tree-based structure of GBDT can make it difficult to extrapolate beyond the training distribution\cite{MESGHALI20241269}.

In this study, we present Tabular Continual Contrastive Learning (TCCL), a method specifically designed to address OOD challenges in tabular data. TCCL adopts a similar contrastive learning framework \cite{Chen2020SimClr, ginanjar2024contrastivefederatedlearningtabular} , which involves an encoder and a header. However, our approach introduces a novel header to handle new data with shifted distributions. 

TCCL consists of three main components: an Encoder, a Decoder, and a Learner Head. The Encoder processes input data into an augmented format while the Decoder translates this encoded data into a new representation. The Learner Head is specifically designed to mitigate the issue of catastrophic forgetting by incorporating a mechanism that acts as a 'break,' effectively preventing the model from losing previously learned information. These features allow TCCL to effectively handle data with shifting distributions. Experimental results demonstrate that TCCL outperforms other models in both classification and regression tasks.

\section{Related Work}
 Neural network models such as Multilayer Perceptron (MLP), Self-Normalizing Neural Networks (SNN), Feature Tokenizer Transformer /FT-Transformer,  Residual Network /ResNet, 
    Deep \& Cross Network  /DCN V2, 
    Automatic Feature Interaction /AutoInt, 
    Neural Oblivious Decision Ensembles / NODE, 
    Tabular Network / TabNet and GrowNet are widely recognized and frequently cited for addressing tabular data prediction problems \cite{Gorishniy2021}. Another approach from the tabular data area is contrastive learning models such as CFL \cite{ginanjar2024contrastivefederatedlearningtabular}, SCARF  \cite{scarf}, and SubTab \cite{subtab}. However, these models are not specifically designed to handle OOD data. In our experiments, we utilized these methods as base models. Note that CFL is not applicable in non-federated learning networks. We are excluding CFL from our experiments.

\section{Problem Formulation}
In machine learning, the goal is to build a model  $f:x\rightarrow y$ that generalizes well to unseen data. A prediction task can be defined as finding a model $f:(.)$ that minimizes the expected error over a dataset  $D_{in}$ with distribution $p_{in}$. This can be expressed in terms of a loss function $\textbf{min } \text{Error}(x,y) _{D} = [L(f(x), y)]$ , which $L$ measures the discrepancy between the model's predictions $f(x)$ and the true labels. A higher loss value indicates poorer model performance  $E\uparrow = P\downarrow$. When a different distribution $D_{ood} \leftarrow  p_{ood}$  is introduced to a model, its performance typically decreases $P(f((x)_{D_{in}})) > P(f((x)_{D_{ood}}))$ \cite{Hsu2020}.

\section{Proposed Method}
We introduce Tabular Contrastive Learning (TCL), an improved approach designed to enhance prediction tasks on tabular data with OOD. 

\subsection{TCCL Main Architecture}
\begin{figure}
    \centering
    \includegraphics[width=0.75\linewidth]{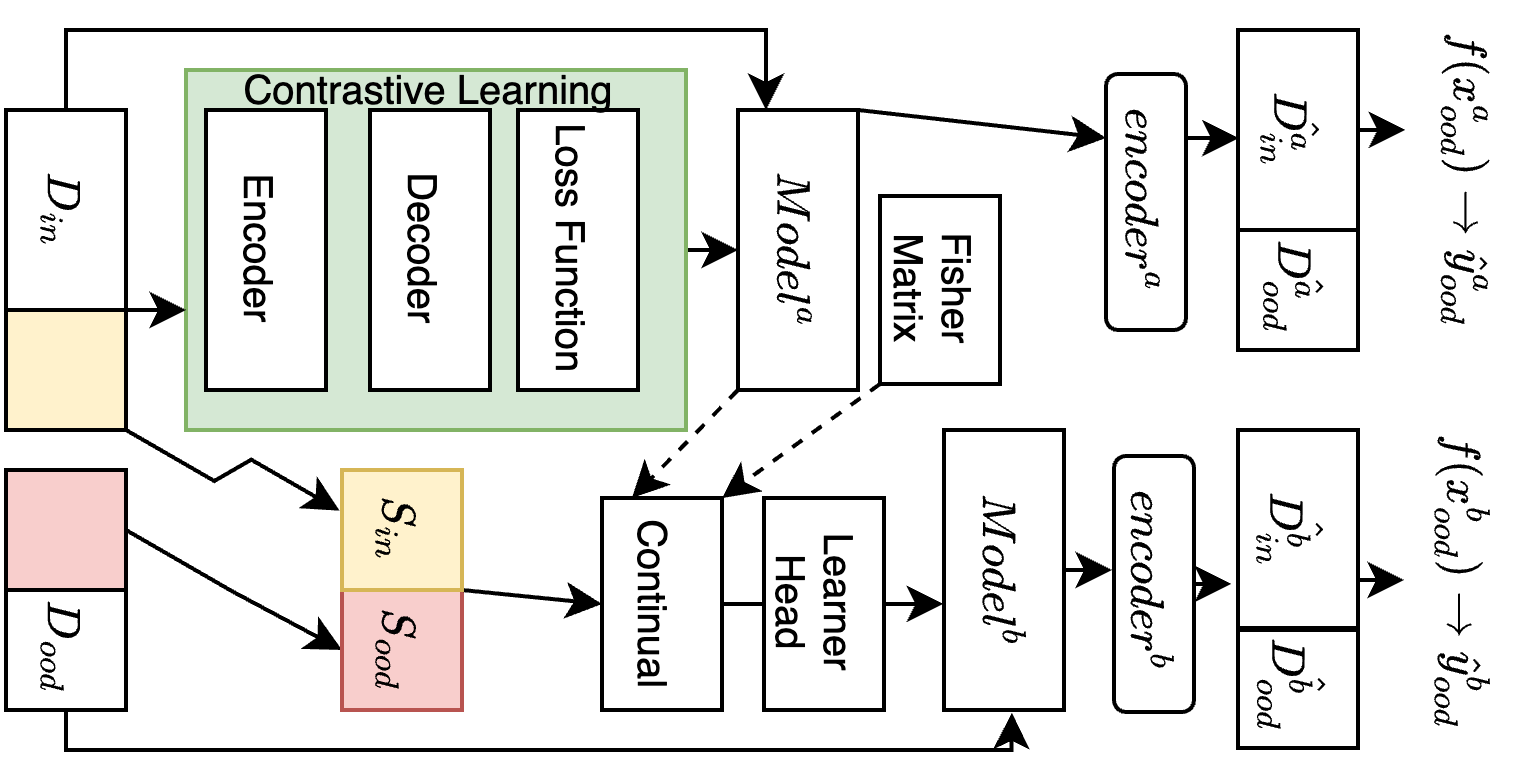}
    \caption{TCCL architecture. TCCL consist of Encoder, Decoder and a Learner Header}
    \label{fig:TCCL_ARC}
\end{figure}

TCCL consist of a tabular contrastive learning model $M^a$, a fisher matrix, a learner header and  an updated continual model $M^b$ , see figure\ref{fig:TCCL_ARC}. TCCL starts with data $D_{in}$ is trained within tabular contrastive learning $M^a$ to produce encoder $E^a$ .  $E^a$ then used to generate new $\hat{D}^a_{in}$ and $\hat{D}^a_{ood} = X^a_{ood}$. Based on the new data, a prediction model $f(x^a_{ood}) \rightarrow \hat{y}^a_{ood}$  is calculated, where $\hat{y}^a_{ood}$ the predicted label from OOD data is based on the model trained from $\hat{D}^a_{in}$. To maintain performance, continual learning is done. $M^a$ is retrained with weight from fisher matrix $F$. $M^a$ is retrained with  $S_{in} \cap D_{in}$ + $S_{ood} \cap  D_{ood}$ . T stop catastrophic forgetting a learner header is used to stop model learning from new data. This is done by setting weight in $M$ near zero during training. The final result is $f(x^b_{ood}) \rightarrow \hat{y}^b_{ood}$ .

\section{Experiment}
\subsection{Datasets}
We utilize 8 diverse tabular datasets and  14 models (including TCL) in our experiment. The datasets are Adult \cite{adult_2}, Helena \cite{helenaJannis}, Jannis \cite{helenaJannis}, Higgs Small \cite{higgssmall}, Aloi \cite{aloi}, Epsilon \cite{epsilon2008}, Cover Type \cite{covtype}, California Housing \cite{ca}, Year \cite{year}, Yahoo \cite{chapelle2011yahoo}, and Microsoft \cite{microsoft}. 

\subsection{OOD Detection}
We have implemented two OOD detection methods, namely OpenMax \cite{Bendale2016} and TemperatureScaling \cite{platt1999probabilistic}.  We separated the OOD data using these detectors to generate two sets $D_{in}$ and $D_{ood}$ where $(D_{in} + D_{ood} = D)$. While $D_{in}$ it is used for training datasets, $D_{ood}$ is used for test datasets. 

\subsection{Prediction ON OOD Dataset}
We experiment with 7 recent models for tabular data e.g. FT-T, DCN2, GrowNet, ResNet, MLP,  AutoInt, and TabR-MLP. In addition, we experimented with the recent implementation of contrastive learning for tabular data SubTab \cite{subtab} and SCARF \cite{scarf}, which comes from a similar domain to our TCL. In addition, we add Lightgbm, CatBoost, and XGB from GBDT models to our base models. 

\section{Result and Evaluation}
\subsection{OOD Detection}
\begin{table}[]
\small
\centering
\caption{OOD experiment settings. Train data is the $D_{in}$,  and test data is the $D_{ood}$.}
\begin{tabular}{>{\raggedright\arraybackslash}p{0.15\linewidth}|>{\raggedright\arraybackslash}p{0.25\linewidth}c|rr}
Dataset            & OOD Detector       & Norm & Train (In) & Test (OOD) \\ \hline
Adult              & OpenMax            & l2   &            31820&            9067\\
Helena             & OpenMax            & l1   &            41724&            13040\\
Aloi               & OpenMax            & l1   &            85982&            522\\
Covtype            & TemperatureScaling& l1   &            464304&            631\\ \hline
California& OpenMax            & l1   &            15665&            1058\\
Year               & OpenMax            & l2   &            370972&            51630\\
Yahoo              & TemperatureScaling& l1   &            473134&            165660\\
Microsoft          & TemperatureScaling& l1   &            957079&            3843
\end{tabular}
\label{tab:ood}
\end{table}
Table \ref{tab:ood} shows the experiment results on splitting the dataset to in distribution $D_{in}$ and OOD data $D_{ood}$.  In the table, we can evaluate the OOD detector and normalisation used during the experiment. Train data is the $D_{in}$  and test data is the $D_{ood}$ . Our experiments are evaluated based on these settings. 

\subsection{Model Performance}
\begin{table}
\small
    \centering
    \caption{Experiment result. F1 score for classification and RMSE for regression. Datasets with (*) mean a regression problem.  Models ($^c$) are contrastive learning based model, and models ($^x$) are GBDT based model.)}
    \begin{tabular}{l|cccc|cccc} \hline 
         &  AD$\uparrow$&  HE$\uparrow$&  AL$\uparrow$& CO$\uparrow$&   CA*$\downarrow$&YE*$\downarrow$&YA*$\downarrow$&MI*$\downarrow$\\ \hline  
         FT-T&  0.782&  0.153&  0.407&  -&   0.867&6.461&-&-\\ 
         DCN2&  0.744&  0.129&  0.414&  0.58&   2.602&7.054&0.645&0.746\\   
 GrowNet& 0.465& -& -& -& 0.969& 7.605& 1.01&0.769\\  
 ResNet& 0.652& 0.10& 0.437& 0.694& 0.892& 6.496& 0.639&0.736\\ 
 MLP& 0.508& 0.146& 0.326& 0.617& 0.894& 6.488& 0.657&0.741\\  
 AutoInt& 0.78& 0.133& 0.401& 0.608& 0.89& 6.673& -&0.739\\
 TabR-MLP& 0.688& 0.165& 0.429& 0.688& 2.677& 2e5& 1.285&0.79\\ \hline
TCCL$^c$& 0.861& \textbf{0.236}& \textbf{0.510}& \textbf{0.972}& \textbf{0.745}&\textbf{6.329}&\textbf{0.636}&\textbf{0.734}\\

 Scarf$^c$& 0.720& 0.00& 0.00& 0.091& -& -& -&-\\
 SubTab$^c$& 0.714& 0.146& 0.322& 0.59&   1.012&6.668&0.656&0.744\\ \hline
 Lightgbm$^x$& 0.591& 0.080& 0.177& 0.219& 0.848& 6.565& 0.661&0.740\\
 CatBoost$^x$& \textbf{0.927}& 0.152& -& 0.753& 0.827& 6.622& 0.655&0.733\\
 XGB$^x$& 0.925& 0.127& 0.328& 0.700& 0.845& 6.867& 0.654&0.739\\
    \end{tabular}
    
    \label{tab:results}
\end{table}
Table \ref{tab:results} shows the results of our experiments. Deep learning models like FT-T, DCN2, ResNet, MLP, AutoInt, and TabR-MLP showed varied performance across datasets, with notable struggles such as GrowNet's poor performance on the Adult dataset (0.465) and DCN2's high RMSE (2.602) on California Housing. TCCL contrastive learning models outperformed competitors SCARF and SubTab, achieving impressive results, including a 0.972 F1 score on Covtype and low RMSE scores on various regression tasks. Among GBDT models, CatBoost and XGBoost performed exceptionally well, particularly on the Adult dataset (F1 scores: 0.927 and 0.925), while LightGBM generally underperformed compared to its GBDT counterparts.

TCCL exhibits superior generalization capabilities in both classification and regression tasks, effectively handling out-of-distribution scenarios in various tabular data contexts. However, traditional GBDT models remain competitive, particularly excelling in specific datasets such as the Adult dataset, where they outperform TCCL. Despite this exception, TCCL consistently demonstrates strong performance across diverse tasks, indicating that its architecture is well-suited to addressing the challenges posed by distribution shifts in tabular data.

\section{Conclusion }
While out-of-distribution presents significant challenges to machine learning, TCCL  demonstrates promising results in handling this type of data. TCCL performs well on most datasets, except for the Adult dataset, where the GBDT model outperforms it. However, in other datasets, TCCL surpasses both deep learning models and GBDT.  GBDT ranks as the second-best model when it comes to out-of-distribution scenarios. Although TCCL shows strong results, future studies should assess its robustness by testing it on a wider range of datasets. 

\begin{footnotesize}
\bibliographystyle{unsrt}
\bibliography{sample}

\end{footnotesize}

\end{document}